\documentclass{article}
\usepackage{ijcai15}

\usepackage{times}
\usepackage{helvet}
\usepackage{courier}
\usepackage[utf8]{inputenc}
\usepackage{amsmath}
\usepackage{dsfont}
\usepackage{graphicx}
\usepackage{color}
\usepackage{caption}
\usepackage{subcaption}
\usepackage{nicefrac}
\usepackage{multirow}
\usepackage{multicol}

\newcommand{\lsym}{\textit}

\newcommand{\fuzzymln}{\textsc{Fuzzy-MLN}}

\newcommand{\wupsim}{\sim_{\sqsubseteq}}%_{\textit{WUP}}}

\newcommand{\isa}{\textit{is-a}}
\newcommand{\hassense}{\textit{instance-of}}
\newcommand{\semrole}{\textit{sem\_role}}

\newcommand{\cfill}{\textit{fill.v.01}}
\newcommand{\cmilk}{\textit{milk.n.01}}
\newcommand{\ccupcont}{\textit{cup.n.01}}
\newcommand{\ccupunit}{\textit{cup.n.02}}

\newcommand{\cadd}{\textit{add.v.01}}

\newcommand{\wfill}{\textit{Fill}}

\newcommand{\wmilk}{\textit{milk}}
\newcommand{\wadd}{\textit{Add}}
\newcommand{\wcup}{\textit{cup}}

\newcommand{\sfill}{\textit{fill-sense}}

\newcommand{\smilk}{\textit{milk-sense}}
\newcommand{\sadd}{\textit{add-sense}}
\newcommand{\scupcont}{\textit{cup-sense}_1}
\newcommand{\scupunit}{\textit{cup-sense}_2}

\newcommand{\actionverb}{\textit{action\_verb}}
\newcommand{\theme}{\textit{theme}}
\newcommand{\goal}{\textit{goal}}
\newcommand{\amount}{\textit{amount}}
\newcommand{\pos}{\textit{has\_pos}}

\begin{document}
% The file aaai.sty is the style file for AAAI Press 
% proceedings, working notes, and technical reports.
%

 \title{Reasoning about Unmodelled Concepts --- \\
   Incorporating Class Taxonomies in Probabilistic Relational Models}

%\author{Submission No. 185}

\author{Daniel Nyga and Michael Beetz\\
Institute for Artificial Intelligence\\
University of Bremen, Germany\\
\texttt{\{nyga, beetz\}@cs.uni-bremen.de}
}
\maketitle
\begin{abstract}

  A key problem in the application of first-order probabilistic
  methods is the enormous size of graphical models they imply. The
  size results from the possible worlds that can be generated by a
  domain of objects and relations. One of the reasons for this
  explosion is that so far the approaches do not sufficiently exploit
  the structure and similarity of possible worlds in order to encode
  the models more compactly.  We propose fuzzy inference in 
  Markov logic networks, which enables the use of taxonomic
  knowledge as a source of imposing structure onto possible
  worlds. We show that by exploiting this structure, probability
  distributions can be represented more compactly and that the
  reasoning systems become capable of reasoning about concepts not
  contained in the probabilistic knowledge base.

\end{abstract}

% % % % % % % % % % % % % % % % % % % % % % % % % % % % % % % % % % % %
% HERE THE CONTENT STARTS
% % % % % % % % % % % % % % % % % % % % % % % % % % % % % % % % % % % %

% % % % % % % % % % % % % % % % % % % % % % % % % % % % % % % % % % % %
% INTRODUCTION
\section{Introduction}

Many real-world reasoning problems require the combination of 
relational representations with inference mechanisms that can solve 
the problems by reasoning from incomplete, ambiguous, inaccurate 
and even contradictory information. Examples of such reasoning 
tasks are the interpretation of natural-language~\cite{beltagy14}, 
object recognition for robot perception~\cite{icra14ensmln} or
intent recognition in human-robot interaction~\cite{sukthankar2014plan}.

First-order probabilistic models \cite{getoor07prm} have great potential to
serve as powerful problem-solving tools for such application domains:
joint probability distributions over the instantiated
relations that describe the possible worlds in the respective domain
can be queried for any aspect $Q$ contained in the model 
given any evidence $E$, $P(Q\mid E)$.

These powerful reasoning capabilities, however, come at the cost of 
computational complexity in learning and reasoning as the size of 
the domain under consideration grows. As a consequence, practical 
applications are mostly bound to small application domains with 
limited complexity. Many knowledge systems, however, have to work 
in open worlds: they are equipped with knowledge bases (KB) that 
have to answer queries about unseen situations that have not been 
accounted in their design, such as the examples mentioned above.

Hence, the application of expressive probabilistic representation 
methods requires the inference mechanisms to support 
\emph{off-domain reasoning} -- reasoning about concepts that are 
not explicitly represented in the KB.  Most of the 
probabilistic models, however, do not support off-domain reasoning. 
They require every symbol subject to reasoning to be explicitly 
represented. On the other hand, learning a probability distribution 
with all possible concepts is hopelessly infeasible.

We therefore aim at developing reasoning mechanisms that are able to 
rapidly yet flexibly generalize and learn from very few examples, 
which has also been identified as key features in human cognition 
\cite{tenenbaum2011grow,bailey1997push}. An obvious idea to tackle
this is to take into account knowledge about the taxonomic 
structure of the reasoning domain, which is captured by ontological 
knowledge representations such as description logics. To this end, 
the correlation between the semantic similarity of concepts, and 
the similarity of their relational structure can be 
exploited for reasoning in probabilistic relational models to 
transfer the learned knowledge to classes unseen in the training 
data. 

We propose \fuzzymln s as a probabilistic reasoning framework for 
Markov logic networks (MLN)~\cite{richardson06ml}. \fuzzymln s 
exploit the semantic similarity of concepts in a taxonomy in order to 
handle off-domain concepts in previously unseen situations in a 
meaningful way and hence allow efficient generalization from very 
sparse data whilst the original representation formalism of MLNs remains unchanged. The key 
idea of \fuzzymln s is to learn joint probability distributions 
\textit{conditioned on} large taxonomic knowledge bases that are 
assumed to be given as factual knowledge. Indeed, a number of 
comprehensive high-quality taxonomies exist that have been 
carefully designed to reflect the semantic similarity of concepts~
\cite{fellbaum98wel,lenat95cyc}, which we use as an 
implementational basis. In contrast to existing probabilistic methods 
incorporating class hierarchies, \fuzzymln s do not target 
reasoning about the taxonomic structure as such. This comes with the advantage 
that the concepts subject to reasoning do not 
need to be exhaustively modelled in the probabilistic KB. 
This enables (1) compact representation of knowledge, (2) 
powerful generalization from sparse training data and (3) reduced
complexity of learning and inference.
In particular, our contributions are the following:
\begin{enumerate}

        \item We present an approach for reasoning about unknown concepts 
           by exploiting semantic similarity to known concepts in Markov logic,
	   which typically impedes a compact representation of
          classes that are hierarchically organized in a taxonomy.

        \item We propose a reasoning framework for MLNs that enables
	  inference in presence of vague evidence, which allows a very 
	  compact representation of knowledge in MLNs and learning from sparse data.

        \item We demonstrate the strengths of \fuzzymln s by the
          example of word-sense disambiguation and showcase its strong
          generalization abilities.
\end{enumerate}

\section{Running Example}
\label{sec:applications}

Let word-sense disambiguation (WSD) and semantic role
labelling (SRL), which are widely studied problems in natural-language
processing, be our running examples. Solving these problems enables
software systems to interpret incomplete and ambiguous instructions
and transform them into well-defined action specifications. 
More specifically, take the terms `cup' and `milk' and their usage in
the two instructions `fill a cup with milk' and `add a cup of
milk'. In the former case, `cup' refers to a drinking mug, a physical
object that can hold milk. In the latter case, it refers to a
measurement unit specifying the amount of milk to be added to
something not further specified.

\begin{figure}[h]
        \includegraphics[width=\columnwidth]{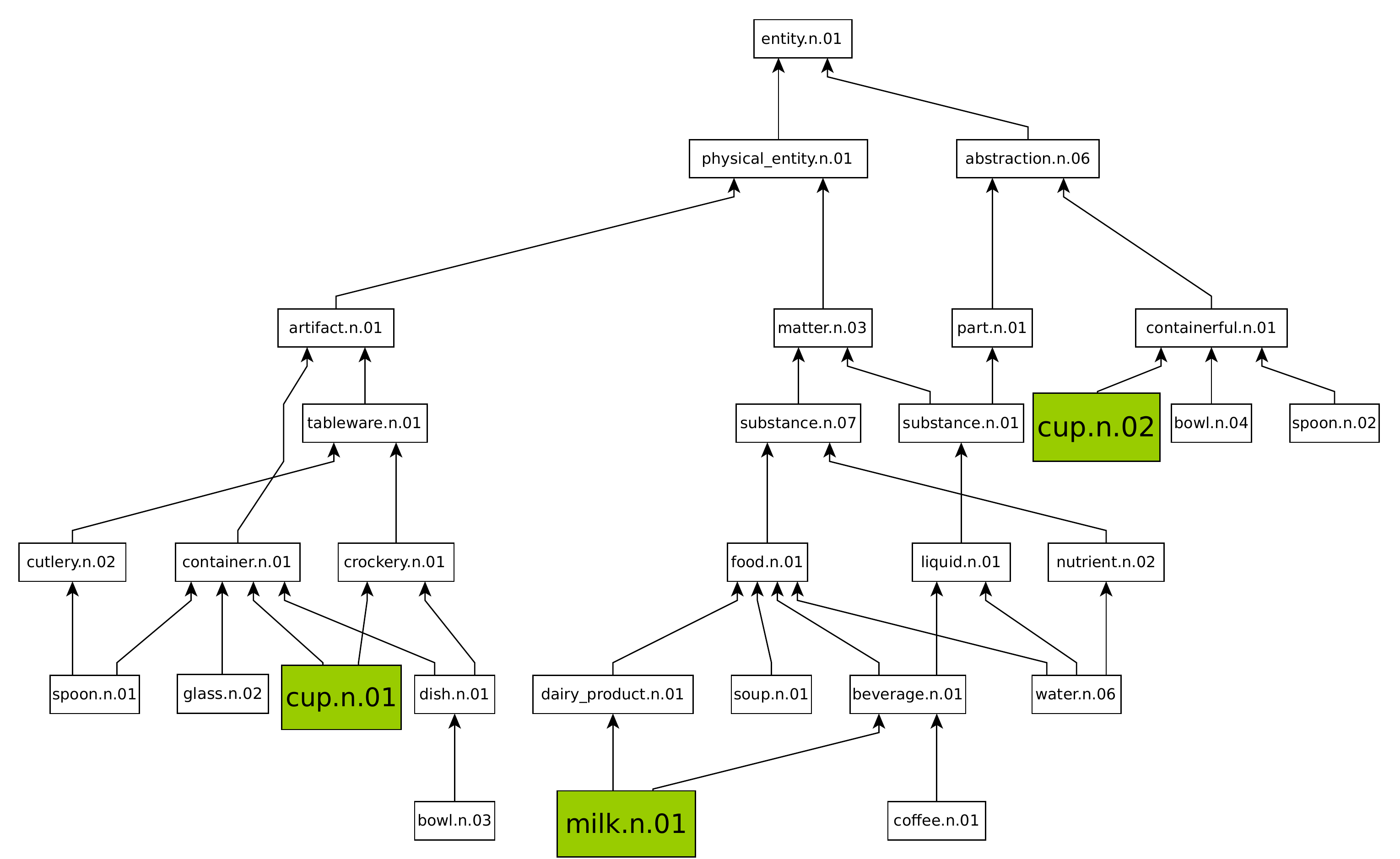}
        \caption{Excerpt of the WordNet taxonomy of concepts for the `containers-\&-liquids' example.}
        \label{fig:taxonomy-example}
\end{figure}

Figure~\ref{fig:taxonomy-example} shows a small excerpt of the WordNet
taxonomy %\footnote{All concept names referred to in this work
  %correspond to concepts of the NLTK toolbox (\texttt
  %{http://www.nltk.org}).} 
  of possible word senses covering this
example. Using the taxonomy we can represent the two instructions
using the following logical assertions 

{\small
\begin{tabular}{l|l}
  \textbf{instruction 1:} & \textbf{instruction 2:} \\
  \hline
  $\hassense(\wfill,\sfill)$ & $\hassense(\wadd,\sadd)$\\
  $\isa(\sfill,\cfill)$ & $\isa(\sadd,\cadd)$\\
  
  $\hassense(\wcup,\scupcont)$ & $\hassense(\wcup,\scupunit)$\\
  $\isa(\scupcont,\ccupcont)$ & $\isa(\scupunit,\ccupunit)$\\
  
  $\hassense(\wmilk,\smilk)$ & $\hassense(\wmilk,\smilk)$\\
  $\isa(\smilk,\cmilk)$ & $\isa(\smilk,\cmilk)$\\
  
  $\semrole(\wcup,\goal)$ & $\semrole(\wcup,\amount)$\\
  $\semrole(\wmilk,\theme)$ & $\semrole(\wmilk,\theme)$,
  
\end{tabular}} 

\noindent The assertions assign a word sense (\hassense) to each
word. The word sense is linked to the taxonomy using the \isa\
predicate. In addition, the predicate \semrole\ states the semantic role that
the word takes in the instruction, whether it is the object acted on,
the source of the stuff to be transferred, the destination, the action
verb, and so on.  

Now suppose we have a taxonomy and two examplary instructions to learn
from: `fill a cup with milk' and `add a cup of milk'. For the sentence
'fill water into the pot' a probabilistic reasoner should infer that
water is the stuff to be added and the pot the destination, even when
'water' and 'pot' are not contained in the probabilistic knowledge
base. The reason is that 'water' is a liquid like 'milk' and therefore
semantically similar and that a 'pot' is also a container and
therefore similar to a cup. Current first-order probabilistic
reasoning frameworks cannot perform this pattern of reasoning as they
are restricted to concepts contained in their probabilistic knowledge
base.

In the following sections we will explain how we can extend MLNs to
perform such reasoning tasks. Note that the reasoning tasks we are
interested in are not whether or not two concepts are similar. This is
already asserted in the taxonomy. We rather want to infer the concepts
that entities belong to and the role they take in actions.

\section{Foundations}
\label{sec:foundations}

Before defining \fuzzymln s we first introduce the formal groundwork
they are based on: \textit{Description Logics} (DL), \textit{Fuzzy
  Logics} (FL) and \textit{Markov logic networks} (MLN).

\paragraph{Markov Logic Networks}
Our basic formalism for representing, learning, and reasoning about 
first-order probabilistic knowledge bases are MLNs.
Formally, an MLN $L$ is given by a set of pairs 
$\langle F_i, w_i \rangle$, where $F_i$ is a formula in first-order 
logic (FOL) and $w_i$ is a real-valued weight. For each finite domain of 
discourse $D$, a ground Markov random field (MRF) can be instantiated 
by introducing to the MRF a Boolean variable for each ground atom 
and a binary feature $\widehat{f}_j : \mathcal{X} \mapsto \{0,1\}$ 
for each ground formula $\widehat{F}_j$, whose value for a possible world $x \in \mathcal{X}$ is 1 
if the respective ground formula is satisfied in  $x$ and 0 
otherwise, and whose weight is $w_j$.
The ground MRF specifies a probability 
distribution over the set of possible worlds $\mathcal{X}$ according to
\begin{align}
	P(X=x) 
		&=\frac{1}{Z}\exp\left(\sum_{j=1}^{|G|} \hat{w}_j \hat{f}_j(x)\right),\label{eq:mln}
		%&=& \frac{1}{Z}\exp\left(\sum_{i=1}^{|L|} {w}_i {n}_i(x)\right),\label{eq:aggr}
%\label{eq:mln:world-prob}
\end{align}

\noindent where $Z$ is a normalization constant and $G$ is an indexed
set of weighted ground formulas, i.e.\ a set of pairs $\langle\widehat{F}_j,\widehat{w}_j\rangle$
containing a pair $\langle\widehat{F}_j,\widehat{w}_j\!\!=\!\!w_i\rangle$ for every ground
formula $\widehat{F}_j$ of the formula $F_i$, and $\widehat{f}_j$
is the feature associated to the $j$-th pair.
 
\paragraph{Description Logics}
\label{sec:description-logics}

The formulas in our probabilistic KBs are not 
independent of each other. Rather there are many constraints 
between them. For example, if an entity $e$ is an instance of 
the concept \textit{Cup} then there might be another entity $e'$
such that the relation $\textit{holds}(\cdot,\cdot)$ holds for the pair 
$\langle e,e' \rangle$, i.e.\ the assertion $\textit{holds}(e,e')$
must hold. DL are appropriate representation 
mechanisms to state such relations. In DL, these constraints are 
asserted as terminological axioms of the form \emph{c} $\doteq$ 
\emph{exp}. In our case, we can assert, for instance, \emph{Cup} $\doteq$ \emph
{Container} $\sqcap$ $\exists$ \emph{holds.Liquid} $\sqcap\ \exists$ \emph{has.Handle} in order to state 
that the concept of a cup is the intersection of the 
concept of a container that has a handle and holds some liquid. 
For the purpose of this work it is important to note that the concept
that is defined `inherits' the constraints from the concepts it is
defined with forming a taxonomy relation $\sqsubseteq$. Therefore, the similarity of the relational structure of
concepts is highly correlated with their distance in the concept
taxonomy. $\top$ denotes the set of all concepts in $\sqsubseteq$.

\paragraph{Semantic similarity} The semantic similarity of two
concepts in DL-based representations can be characterized in terms of
the relative location of the two concepts in the taxonomy.  Popular
measures take into account the lengths of the shortest paths between
two concepts in the respective taxonomy graph. The shorter the paths
connecting the two nodes in the graph are, the more similar the
respective concepts are assumed to be.  Among those similarity
measures, the WUP similarity~\cite {wu94verbs}
$\wupsim:\top\times\top\mapsto[0,1]$ is the
most widely used. It defines the semantic similarity on concepts in a
class taxonomy as $c_1\!\wupsim\! c_2:=\frac{2\cdot
  \textit{depth}(\textit{lcs}(c_1,c_2))}{\textit{depth}(c_1)+\textit{depth}(c_2)},$
where $lcs(\cdot,\cdot)$ denotes the least common super-concept of two
concepts in $\sqsubseteq$.

\paragraph{Fuzzy Logic}
\label{sec:fuzzy-logic}

As we want to reason about concepts that are not contained in our 
probabilistic model, we need representational means to express our 
expectations about the properties of an unknown concept, which we 
are uncertain of. To do this, we intend to replace the binary truth 
values in MLNs with degrees of beliefs about whether or not 
relations hold for a concept not contained in the probabilistic 
model. We use \textit{fuzzy logic} (FL) for this purpose, a 
multi-valued extension of propositional logic (PL). FL has its 
foundations in the theory of fuzzy sets, in which elements belong 
to a set only to a certain degree. Formally, a fuzzy subset $x$ of 
a set $X$ is a pair $\left<X,\pi_x\right>$, where $X$ is called the 
\textit{universe} and $\pi_x:X\mapsto[0,1]$ determines the 
degree to which a particular element actually belongs to $x$, which 
is called the \textit {membership function}. In FL, the universe 
$X$ is given by the set of atomic propositions and $\pi_x$ is a 
fuzzy interpretation of $X$ assigning every proposition in $X$ a 
real-valued degree of truth.  It provides a calculus analogous to 
the calculus of PL: If $A$ and $B$ are 
propositions in FL, then the logical connectors with respect to $x$ 
are defined as $A\land B \!:=\!\min\left(\pi_x(A),\pi_x(B)\right)$, 
$A\lor B\!:=\! \max\left(\pi_x(A),\pi_x(B)\right)$, and $\lnot A 
\!:=\! 1-\pi_x(A)$. Note that the multi-valued logical calculus of FL 
reduces to its binary counterpart of PL in the extreme cases where 
all propositions have boolean truth values.

\section{\fuzzymln s}
\label{sec:fuzzy-mlns}

\begin{it}
    
A \fuzzymln\ $F$ is a pair $\langle L, \sqsubseteq\rangle$, where 
$L$ is an MLN and $\sqsubseteq$ is a taxonomy of 
concepts, such that $L$ represents a conditional probability 
distribution 
\begin{align}
    P(\hassense(\cdot,\cdot),\ldots\mid\lsym{\isa}(\cdot,\cdot),\ldots).\label{eq:fuzzymln}
\end{align} In addition, the following conditions hold: 
\begin{enumerate}

\item an entity $e$ in the domain of discourse $D$ is connected to a 
  concept $c$ in the taxonomy $\sqsubseteq$ always by a proposition
  \hspace*{4ex}$\hassense(e,s)\land\isa(s,c),\mbox{ where }s,c\in\top,$

\item all ground atoms of the form $\isa(s,c)$, where $s, c\in\top$ 
  take real-valued degrees of truth $\in[0,1]$, which we call 
  \emph{semantic similarity}. Ground atoms of all other predicates
  take strictly binary truth values $\in\{0,1\}$.

\item The set $\mathcal{X}$ of \emph{possible worlds} represented by $F$ is
  the set of all fuzzy subsets of all ground atoms $X$, where the 
  membership functions for every ground atom \isa(s,c)
  is equal across all possible worlds and is defined as the semantic
  similarity of $s$ and $c$ with respect to $\sqsubseteq$, i.e.
  for all $x\in\mathcal{X}$ and for all $s,c\in\top$:
  $ \pi_x(\isa(s,c)\!)\!=\!s\wupsim c.$

%\item the \emph{semantic similarity} of $\isa(s,c)$ is a
  %function of the WUP distance $\wupsim$ of $s$ and $c$ in the taxonomy:
  %$\isa(s,c):=s\wupsim c$.
  %$\sqsubseteq$: We define $S=\top$ 
  %and the degree to which an entity $s\in\top$ belongs to some 
  %category $c$ as $\pi(s,c):= \textit{sim}_{\textit{WUP}}(s, c)=:\isa(s,c)$. 

\end{enumerate}
\end{it}
In the following, we motivate this definition in more detail. 

\begin{figure*}[t!]
	%\begin{subfigure}[b]{.5\textwidth}
		\includegraphics[width=\columnwidth]{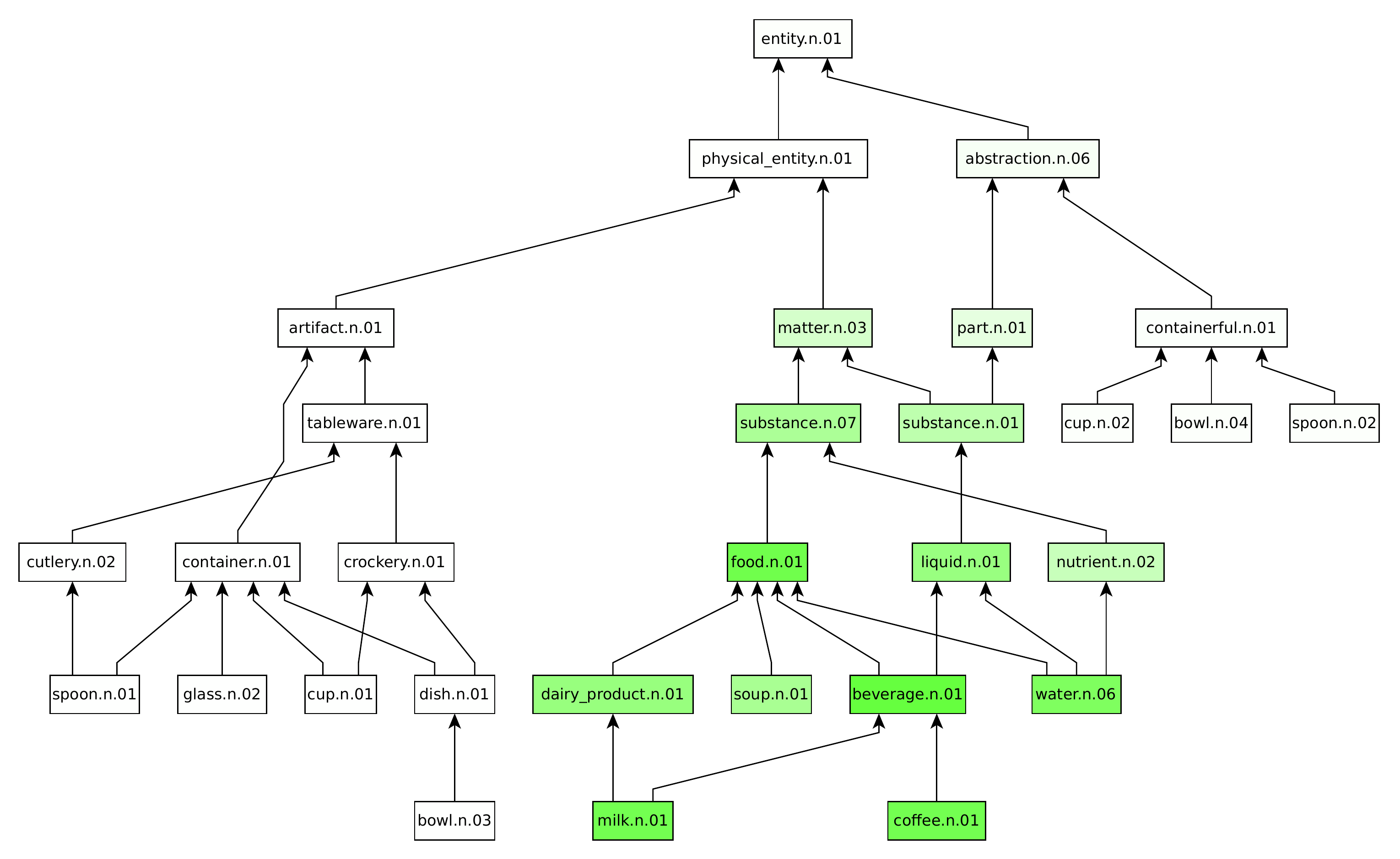}
		%\caption{Posterior of Eq. (\ref{eq:filling-roles}) for $w_1$ given $\semrole(w_1,\theme)$}
		%\label{fig:filling-theme}
	%\end{subfigure}
	%\begin{subfigure}[b]{.5\textwidth}
		\includegraphics[width=\columnwidth]{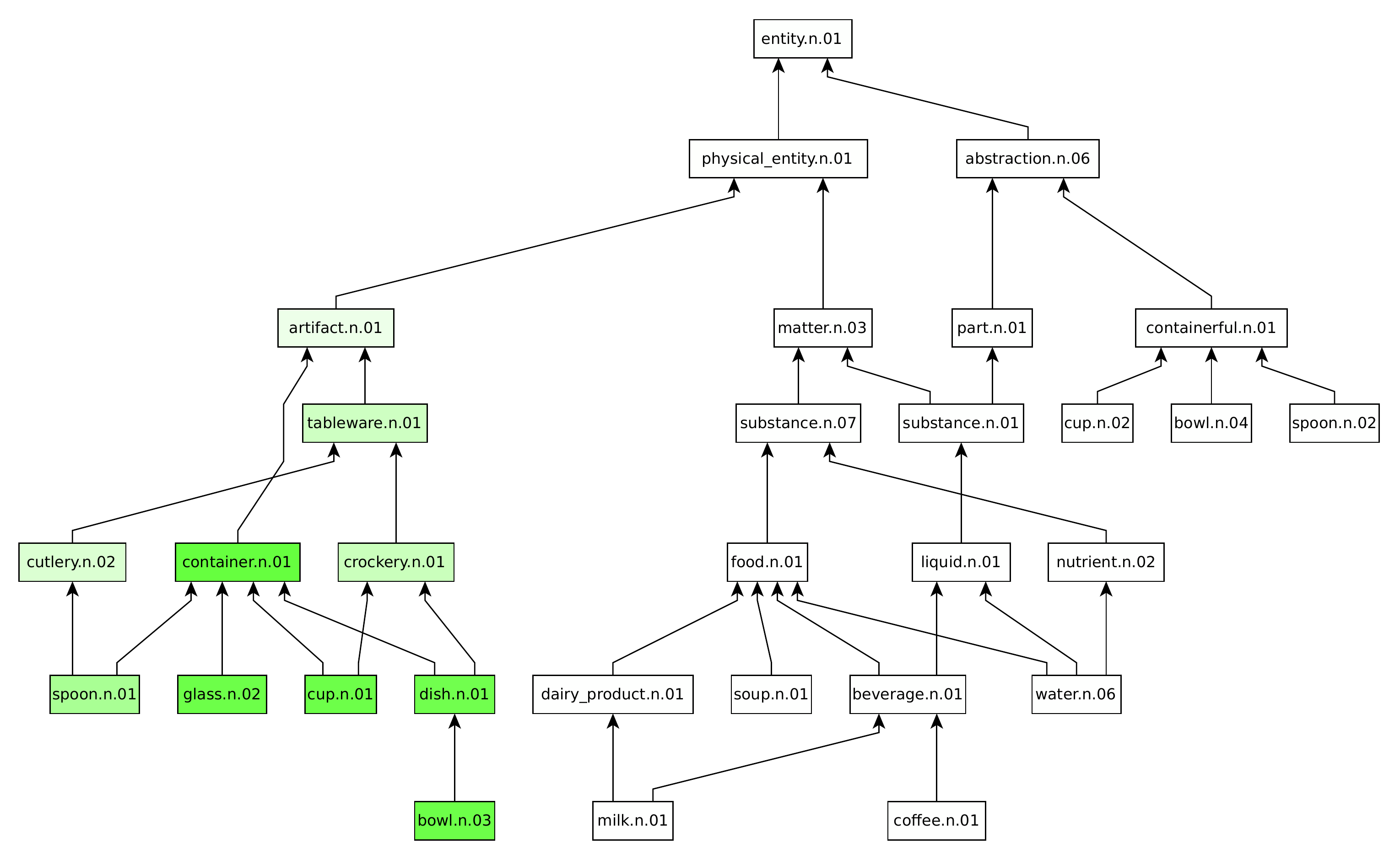}
		%\caption{Posterior of Eq. (\ref{eq:filling-roles}) for $w_2$ given $\semrole(w_2,\goal)$}
		%\label{fig:filling-goal}
	%\end{subfigure}
	%\begin{subfigure}[b]{.5\textwidth}
		%\includegraphics[width=\columnwidth]{knowledge/semantic-core/taxonomy-cluster-adding-bowl-amount}
		%\caption{Posterior of the meaning of `bowl' given an `adding' action.}
		%\label{fig:adding-amount-bowl}
	%\end{subfigure}
	%\begin{subfigure}[b]{.5\textwidth}
		%\includegraphics[width=\columnwidth]{knowledge/semantic-core/taxonomy-cluster-filling-bowl-goal}
		%\caption{Posterior of the meaning of `bowl' given a `filling'.}
		%\label{fig:filling-goal-bowl}
	%\end{subfigure}
	\caption{Posteriors distributions over the taxonomy conditioned
	on semantic roles of a filling action according to Eq. (\ref{eq:filling-roles}). More intense node colors indicate higher probability. \textit{Left:} $w_1$ given $\semrole(w_1,\theme)$
	\textit{Right:} $w_2$ given $\semrole(w_2,\goal)$.}
	%Also for word meanings that are unknown to the MLN, all posteriors represent reasonably well the appropriate word meanings.}
	\label{fig:queries}
\end{figure*}

\paragraph{Probabilistic Semantics}

According to the second condition in our definition, the semantics 
of \fuzzymln s differs from the original in Equation (\ref 
{eq:mln}) in two aspects: First, a possible world $x$ is no longer 
a strictly binary vector assigning a truth value to every ground 
atom but also allows for real-valued degrees of truth. The ground 
MRF of a \fuzzymln\ thus contains binary \textit{and} numerical 
random variables; a real-valued variable for every ground atom of 
the form $\isa(\cdot,\cdot)$ and a binary one for every other 
ground atom. Second, as a consequence, the semantics of the binary 
logical features $\hat{f}_j:\mathcal{X}\mapsto \{0,1\}$ in the 
ground MRF is not applicable any more. We therefore define the 
features associated to every ground formula $\widehat{F}_j$ in the 
MRF to take the form $\hat{f}_j:\mathcal{X}\mapsto[0,1]$, where 
each feature $\hat{f}_j(x)$ evaluates to the truth value 
of its ground formula $\widehat{F}_j$ in $x$ by applying the 
fuzzy logic calculus as described above, i.e.\ 
$\widehat{f}_j(x)=\pi_x(\widehat{F}_j)$. Hence the distribution of 
$F$ becomes 

\begin{align}
    P(X=x) &= \frac{1}{Z}\exp\left(\sum_{j=1}^{|G|} \hat{w}_j \pi_x(\widehat{F}_j)\right).\label{eq:fuzzy-dist}
\end{align}

Condition no.\ 3 in our definition ensures that the probability 
distribution in~(\ref{eq:fuzzy-dist}) corresponds to the 
conditional distribution in (\ref{eq:fuzzymln}): since the truth 
value of a ground atom of the \isa\ predicate is required to 
be equal across all possible worlds $x$, the distribution $P(X\!\!=\!\!x)$
in~(\ref{eq:fuzzy-dist}) is effectively conditioned
on every atom of the form $\isa(\cdot,\cdot)$.

%For the equality (`=') operator in FOL rather having the 
%semantics of identity of two entities instead of equality, we leave 
%its semantics at its original definition, i.e. a term $y=z$ 
%evaluates to 1 iff $y$ is identical to $z$, and 0 otherwise. 

%
%Since the FOL semantics of classical MLNs do not account for 
%real-valued truth values, the calculus of the logical rules in the 
%MLN , which are being transformed to clique potential functions in 
%the ground MRF during the grounding process needs to be adapted 
%appropriately. To do so, it is straighforward to make the step from 
%strictly binary FOL to a multi-valued logical calculus that reduces 
%to its binary counterpart in the extreme cases where all predicates 
%have binary truth values. Consequently, we can require every 
%feature function in the ground MRF to take the form $f_j: 
%\mathcal{X} \rightarrow [0,1]$ and taking as a basis a fuzzy logic 
%semantics. 

A \fuzzymln\ contains two dedicated predicates, \hassense\ and 
\isa, which provide means to incorporate knowledge from the class 
taxonomy into the probabilistic model. In short, \isa\ encodes the 
taxonomic knowledge and \hassense\ is used for expressing 
uncertainty about which categories entities belong to.  By 
differentiating between the two predicates it can be modelled that one 
is certain about the taxonomic structure of the domain 
subject to reasoning but possibly uncertain about which 
concept an entity belongs to. 

In contrast to MLNs, \fuzzymln s do not require all predicates to 
be boolean. Variables (ground atoms) of the form $\isa(s,c)$ in the 
ground MRF take real-valued degrees of truth $\in[0,1]$, which 
express the degree to which $s$ is similar to $c$. Here, a value of 
1 denotes maximal similarity, whereas 0 denotes maximal 
dissimilarity. This allows to represent entities that belong to 
concepts not contained in the probabilistic knowledge base by 
referring to them in terms of their similarity to known 
concepts. Note that in \fuzzymln s, the semantic similarities do 
not have to be computed by probabilistic inference as in other 
formalisms such as PSL. Rather, they are always given by the 
taxonomy structure and exclusively appear as evidence. This makes 
the representation of the conditional distribution in (\ref{eq:fuzzymln}) 
very compact, since the taxonomy structure may be 
collapsed into single numeric values, which scale the contribution 
of every single ground formula to the probability mass (\ref
{eq:fuzzy-dist}) by the similarities of its constituents to concepts that are 
contained in the model. This allows to generalize the learned
knowledge also to classes unseen in the learning data.
In addition, realizing \fuzzymln s without 
having to equip them with the capability of reasoning about 
the similarity relation $\wupsim$ as such, enables us to escape a complexity 
monster. Without making this restriction, inference and learning 
would require us to compute integrals over those variables, 
rendering computational complexity infeasible for practical 
applications. 

Since the distribution of a \fuzzymln\ is conditioned on the 
taxonomic structure of the domain, the second predicate, $\hassense$,
is used to link any entity in the domain of discourse to a concept 
in $\sqsubseteq$. Unlike \isa, \hassense\ is boolean and may be 
subject to inference. Propositions about class memberships of an 
entity $e$ are made in the form $\hassense(e,s) \land 
\isa(s,c)$.

\begin{figure*}[t]
\centering
    \begin{tabular}{ccc}
    \vspace{-1ex}
	%\begin{subfigure}[b]{.33\textwidth}
		\includegraphics[width=.3\textwidth]{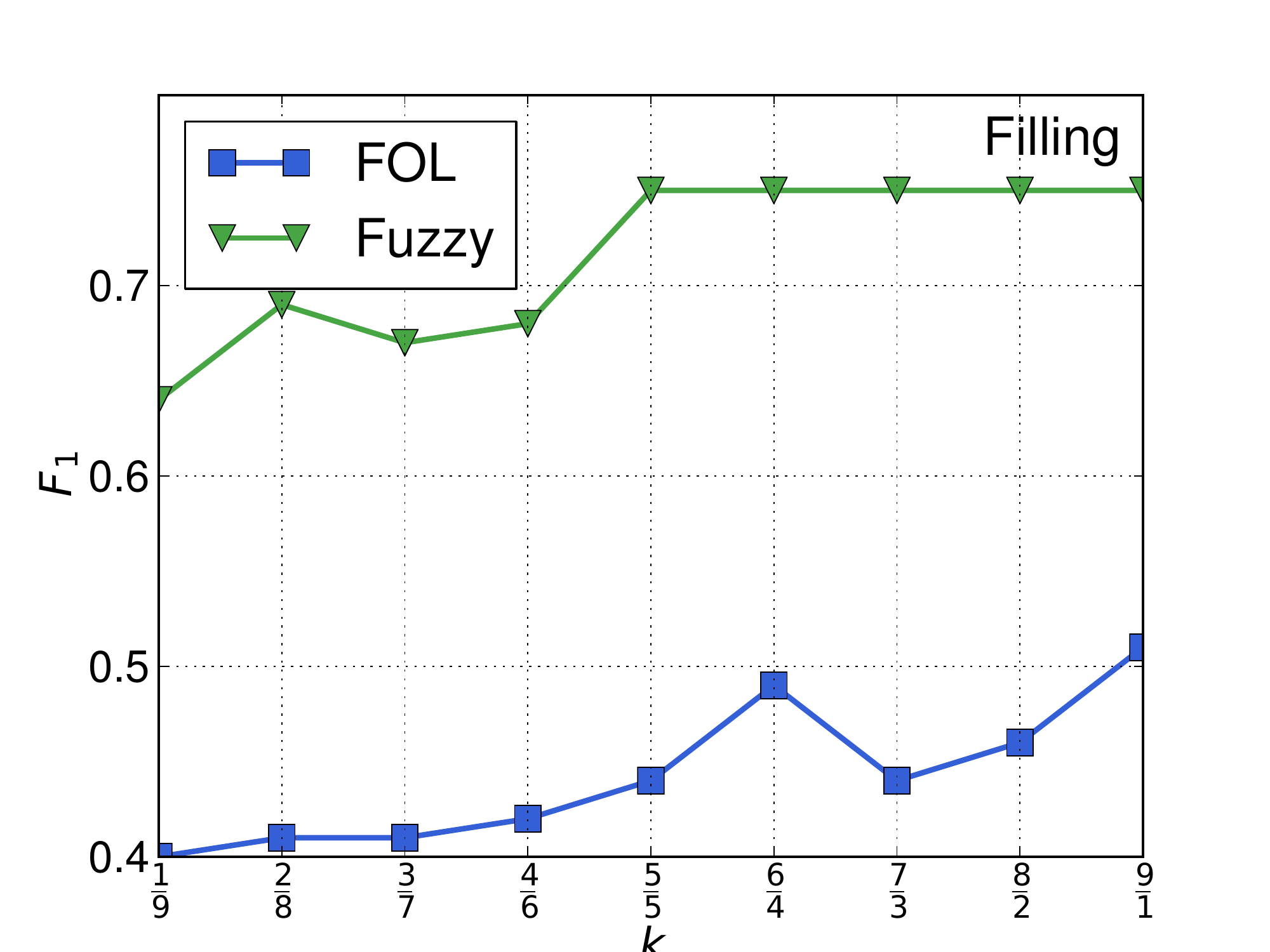} &
		%\caption{Filling (\textit{fill.v.01})}
	%\end{subfigure}
		%\label{fig:fol-vs-fuzzy-filling}
	%\begin{subfigure}[b]{.33\textwidth}
		\includegraphics[width=.3\textwidth]{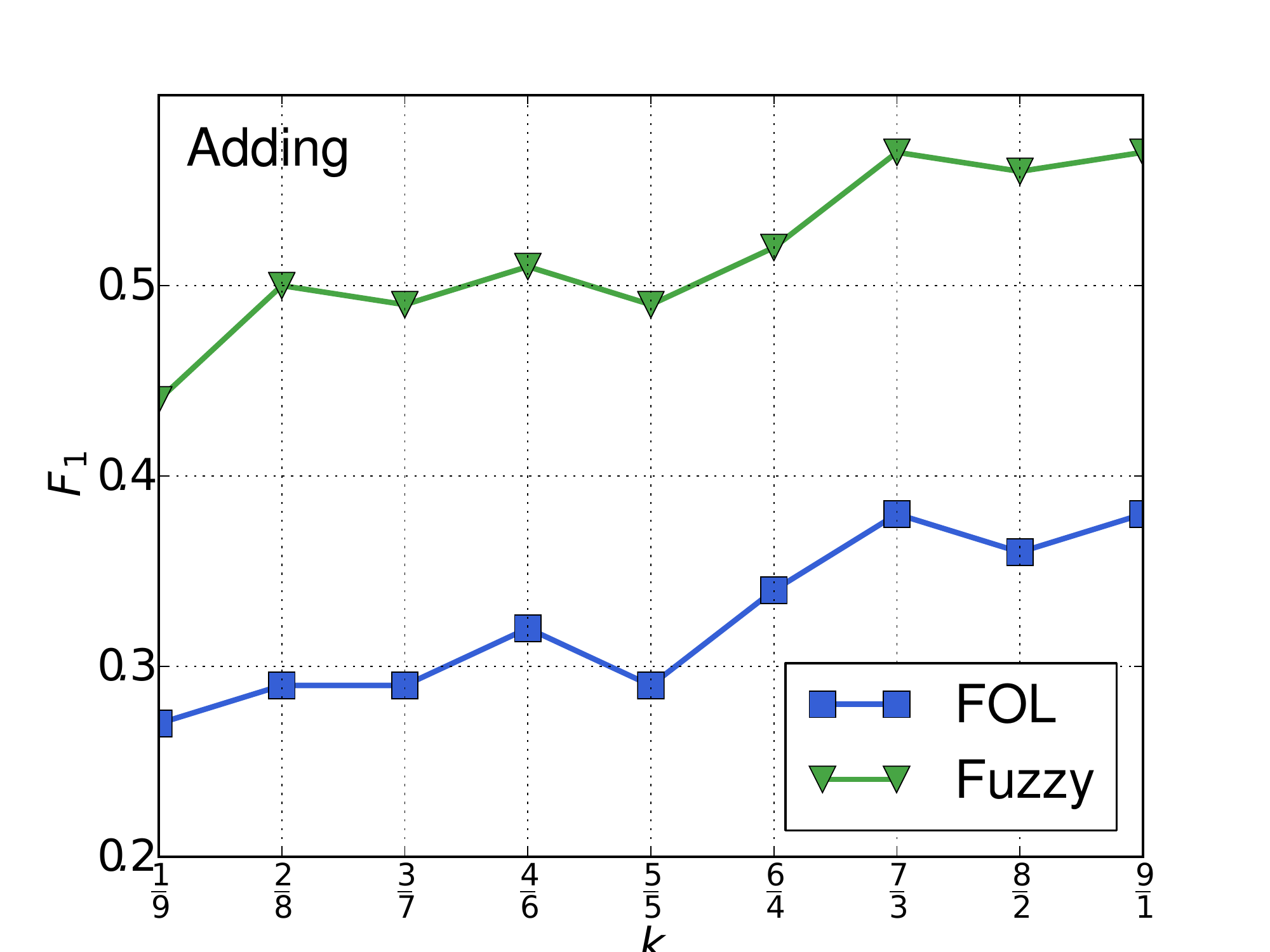} &
		%\caption{Adding (\textit{add.v.01})}
		%\label{fig:fol-vs-fuzzy-add}
	%\end{subfigure}
	%\begin{subfigure}[b]{.33\textwidth}
		\includegraphics[width=.3\textwidth]{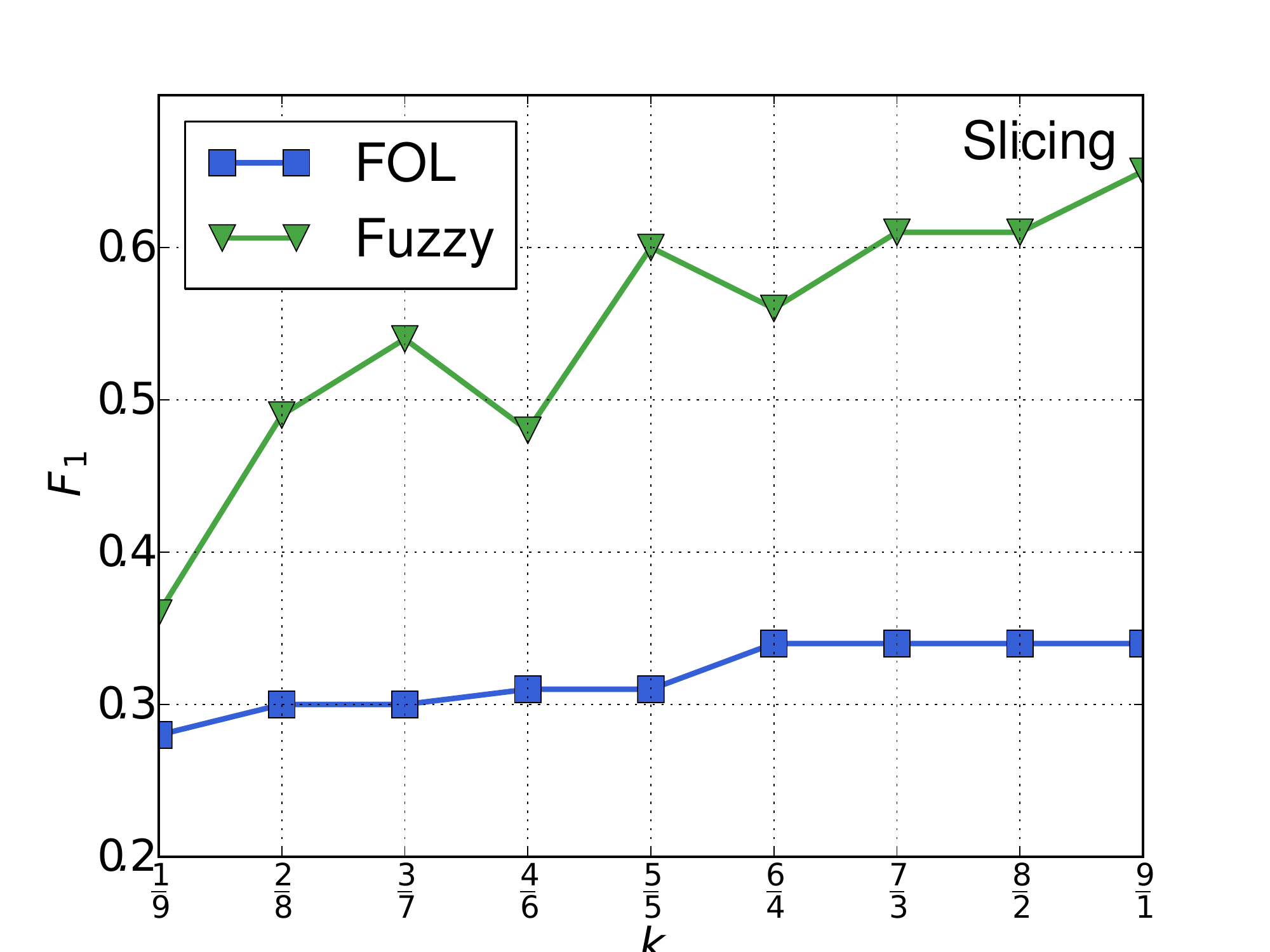}\\
		%\caption{Slicing (\textit{slice.v.03})}
		%\label{fig:fol-vs-fuzzy-slice}
	%\end{subfigure}
        	%\begin{subfigure}[b]{.33\textwidth}
		\includegraphics[width=.3\textwidth]{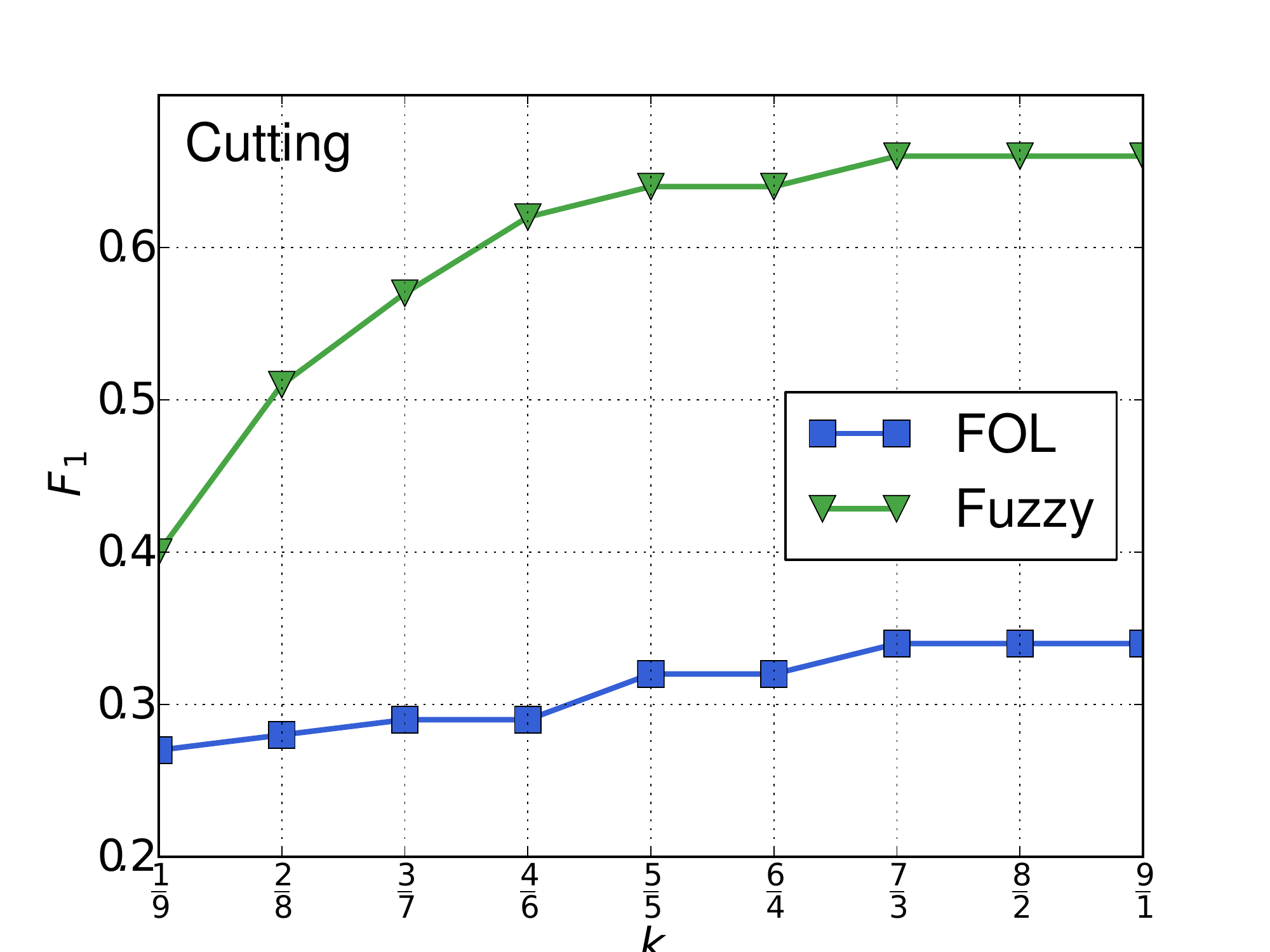}&
		%\caption{Cutting (\textit{cut.v.01})}
		%\label{fig:fol-vs-fuzzy-cutting}
	%\end{subfigure}
	%\begin{subfigure}[b]{.33\textwidth}
		\includegraphics[width=.3\textwidth]{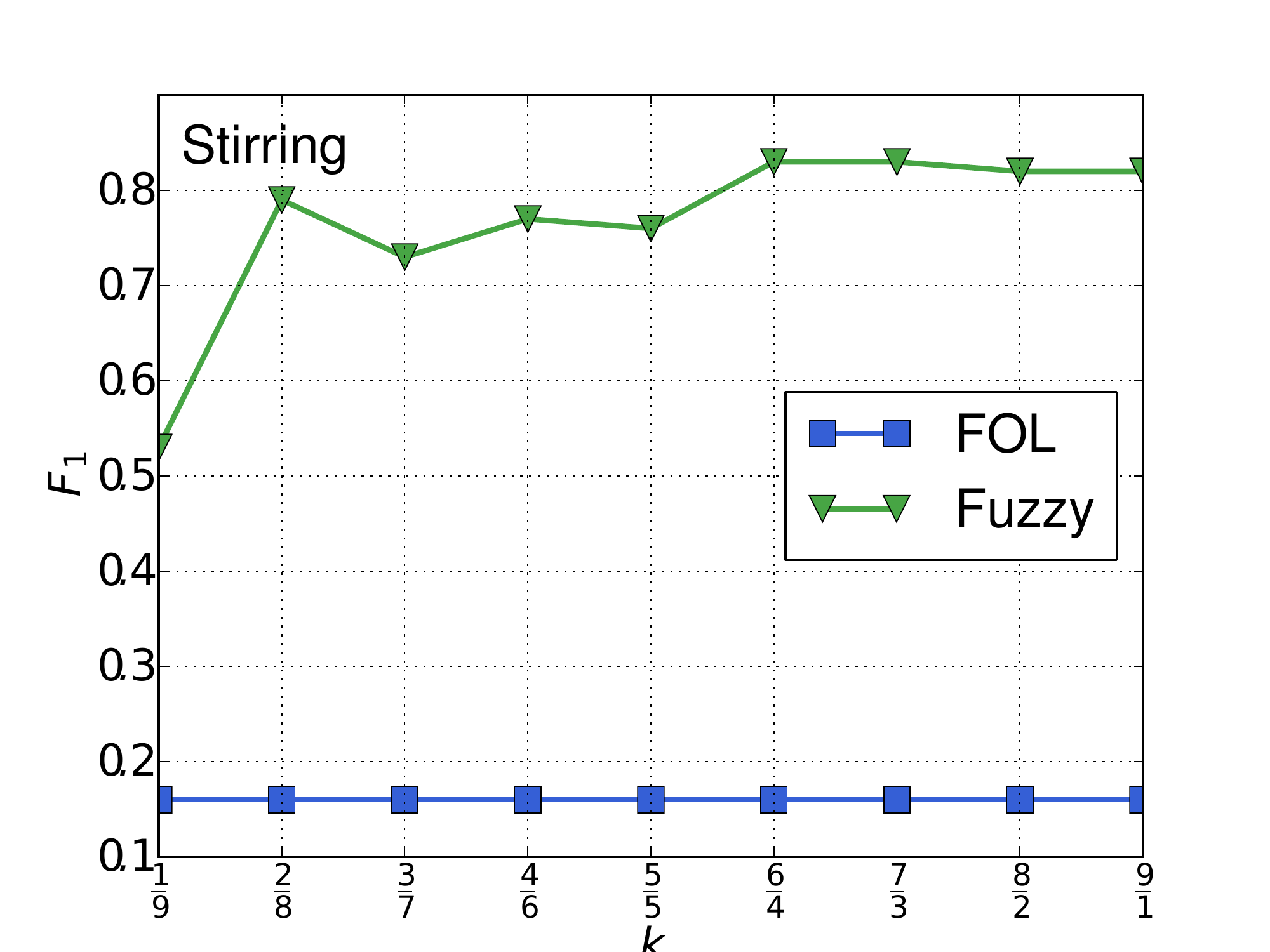}&
		%\caption{Stirring (\textit{stir.v.01/08})}
		%\label{fig:fol-vs-fuzzy-stirring}
	%\end{subfigure}
	%\begin{subfigure}[b]{.33\textwidth}
		\includegraphics[width=.3\textwidth]{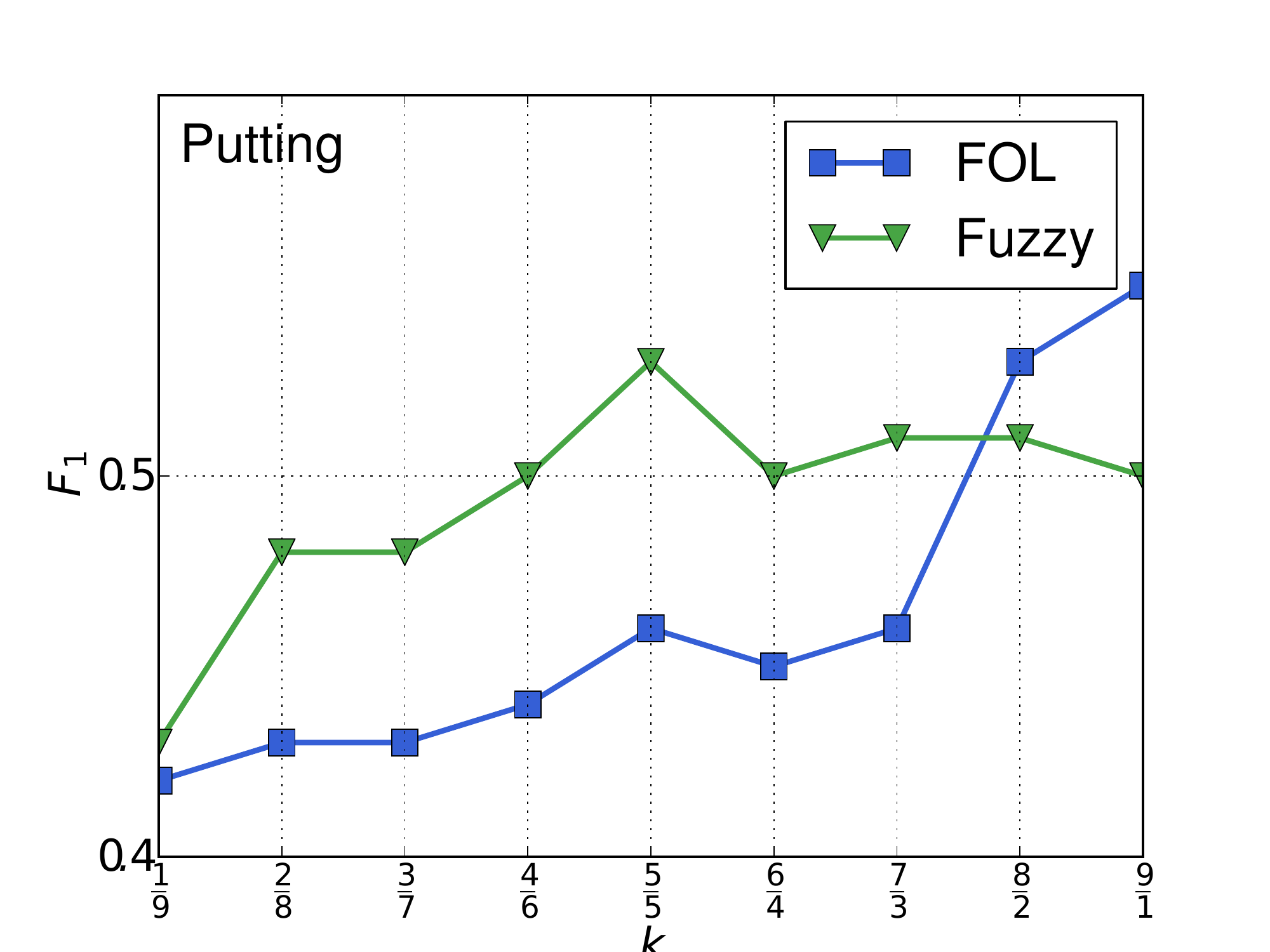}
		%\caption{Putting (\textit{put.v.01})}
		%\label{fig:fol-vs-fuzzy-putting}
	%\end{subfigure}
    \end{tabular}
    \vspace{-1ex}
	\caption{$F_1$ scores for inverse $k$-fold cross validation for $k=\nicefrac{1}{9}\ldots\nicefrac{9}{1}$ using classical MLNs with FOL semantics
        and \fuzzymln s applied to a WSD problem of 20 examples per action verb.}
	\label{fig:fol-vs-fuzzy}
\end{figure*}

Let a minimalistic example illustrate how inference about unknown concepts can be achieved
in \fuzzymln s: Suppose we want to represent the conditional distributions that
parrots can fly and that mammals can not. In Markov logic, we can establish
these distributions in an MLN with, for example, the two weighted formulas
\vspace{-.2ex}
{\footnotesize \begin{align*}
    w_1=\ln(0.9/0.1)\hspace{2ex} &  \textit{flies}(e)\land \hassense(e,\textit{parrot.n.01})\\
		   & \hspace{2ex}\land \isa(\textit{parrot.n.01}, \textit{parrot.n.01})\\
    w_2=\ln(0.1/0.9)\hspace{2ex} & \textit{flies}(e)\land \hassense(e,\textit{mammal.n.01})\\
		   & \hspace{2ex}\land \isa(\textit{mammal.n.01}, \textit{mammal.n.01}).
\end{align*}}%
\noindent In classical MLNs, reasoning can only be performed about instances
of either of the concepts \textit{parrot.n.01} and \textit{mammal.n.01}
because for any other concept, none of the formulas is applicable.
Using a \fuzzymln\ with the same model structure and an underlying 
taxonomy $\sqsubseteq$, however, we can tackle reasoning tasks outside
the model domain, such as
%{\footnotesize
$P(\textit{flies}(\textit{Fred})\,|\,\hassense(\textit{Fred},\textit{turkey.n.01})).$
In this example, there are two ground atoms of the \isa\ predicate,
\isa(\textit{turkey.n.01}, \textit{parrot.n.01}) and \isa(\textit{turkey.n.01}, \textit{mammal.n.01}),
which are, for instance, assigned the truth values 
{\footnotesize
\begin{align*}
\pi_x(\isa(\textit{turkey.n.01},\textit{parrot.n.01})) &= 0.90\\
\pi_x(\isa(\textit{turkey.n.01},\textit{mammal.n.01})) &= 0.01
\end{align*}}%
in every possible world $x$ according to a similarity~$\wupsim$.
Consequently, the influence of the two ground formulas 
{\footnotesize
\begin{align*}
    \widehat{F}_1&=\textit{flies}(\textit{Fred})\land \hassense(\textit{Fred},\textit{turkey.n.01})\\
		   & \hspace*{2ex}\land \isa(\textit{turkey.n.01}, \textit{parrot.n.01})\\
    \widehat{F}_2&=\textit{flies}(\textit{Fred})\land \hassense(\textit{Fred},\textit{turkey.n.01})\\
		   & \hspace*{2ex}\land \isa(\textit{turkey.n.01}, \textit{mammal.n.01})
\end{align*}}%
\noindent on the distribution in (\ref{eq:fuzzy-dist}) is scaled down by 
the similarity of concepts. In the extreme case, where there is 
maximal dissimilarity of two concepts, the contribution of every
ground formula vanishes resulting in a uniform distribution. This is
reasonable since we cannot make any well-informed statement about entities
that are maximally dissimilar to everything that is contained in the
model.

% To this end, we introduce a second dedicated, \textit {binary}
% predicate $\hassense(e,s),$ which connects an entity $e$ with its
% respective sense $s$ (here, the term binary refers to the number of
% truth values the predicate can take; not to the number of its
% arguments). Propositions about different meanings of an entity can
% then be made in the form \hassense(e,s) $\land$ \isa(s,c).

% \begin{eqnarray}
% 	&&\hassense(e,s)\land\isa(s,c),\nonumber
% \end{eqnarray}

%where $e$ is the entity under consideration, $s$ represents its 
%meaning and $c\in\top$ is a class concept in the taxonomy. Using 
%this representational trick of splitting concept assignments
%into two separate predicates allows us to effectively reason about
%concepts that are not explicitly represented in the probabilistic
%model. 

%
%\paragraph{Learning \& Inference}
%
%Since in \fuzzymln s real-valued atoms are restricted to occur as 
%evidence only, there is no need for significant adaptations of 
%existing learning and inference algorithms. In fact, our experience 
%has shown that \fuzzymln s achieve their best performance when 
%being learnt with classical FOL semantics using a discriminative 
%learning procedure~\cite{Anguelov05discriminativelearning,poon06} 
%and to apply the fuzzy semantics for inference only. A possible reason 
%for this is that during learning, it is beneficial to not blur the 
%sharp boundaries between concepts to maintain their discriminative 
%expressiveness.

\paragraph{Running example continued}
Let us now continue with our running example and explain how \fuzzymln
s solve the respective reasoning tasks. 
We consider again the two training databases corresponding to the
instructions (1) `fill a glass with milk' and (2) `add a cup of milk'.
In order to model word sense and role/sense
co-occurrences, we construct a \fuzzymln\ consisting of one single
weighted template formula,
%\footnote{In MLNs, a template variable 
%prefixed by `+' will be expanded to one separate formula for each combination of values
%in the respective variable domains.}:
%For representing the semantic role of a word, we add to our 
%exemplary \fuzzymln a new predicate $\semrole(\word,\role)$, which 
%assigns each word in a sentence a particular semantic a role. As an 
%underlying hierarchy of word meanings, we apply the \textit 
%{WordNet} taxonomy, which we cut down to a small subset for the 
%sake of simplicity and readability. We are discerning four semantic 
%roles, the \actionverb, the \goal\ and \theme\ of some action as 
%well as the \amount\ of the \theme.  
{\footnotesize
\begin{align*}
	&\phantom{\land\ }\hassense(w_1,s_1)\land \isa(s_1,+c_1)\\
	&\land\ \hassense(w_2,s_2)\land \isa(s_2,+c_2)\\
	&\land\ \semrole(w_1,+r_1)\land\ \semrole(w_2,+r_2)\ \land\ w_1\neq\ w_2,
\end{align*} }%
which has been trained with the two databases introduced at the beginning.
%\noindent For performing joint WSD and SRL, an evidence database 
%can be constructed as follows. For every word $w$ in a sentence 
%under consideration, we obtain from WordNet the indexed set of 
%meanings (`\textit{synsets}') $S_w$ the respective word 
%can be assigned to. We then introduce a constant $s_{w_i}$ for every 
%word sense $i\in S_w$ and assert the vague evidences according to 
%their similarity given by the taxonomy 
%$\left<\textit{sim}_{\textit{WUP}}(i,c), \isa(s_{w_i}, c)\right>$
%$$\left<\textit{sim}_{\textit{WUP}}(i,c), \isa(s_{w_i}, c)\right>$$ 
%for all symbols $c$ in the concept domain of the MLN, 
%where a tuple $\left<t,a\right>$ denotes that the ground atom $a$ 
%is assigned a truth value of $t$.

%Note that, for the similarities 
%of word meanings, we only need to take into account the concepts 
%that are already contained in the learnt model, i.e. solely the 
%concepts that are present in the training data. This makes the 
%proposed model a very concise representation of concept 
%correlations since super-classes in the taxonomy do not need to be 
%represented explicitly. The taxonomy is rather collapsed to the 
%conceptual similarities and reasoning about new situations 
%corresponds to matching them against the patterns that have been 
%seen during training. Thus the model can perform reasoning about
%\textit{any} concept of the WordNet taxonomy.

In order to illustrate that the learned MLN indeed reasonably 
generalizes across classes, we visualize the posterior 
distributions over the WordNet taxonomy for two exemplary queries. 
Figure~\ref{fig:queries}
shows the posteriors of two queries 
for the meaning of a word representing the \theme\ of a `filling' 
activity and its \goal, respectively, i.e.
{\footnotesize
\begin{equation}
	\arraycolsep=1pt
	\begin{array}{ll}P\left(\begin{array}{ll}
				\begin{array}{l}
					\hassense(w_1,s_1),\\
	                \hassense(w_2,s_2)
	            \end{array}
	        \end{array}
	                \right| &
	\left.\begin{array}{l}
	\hassense(w',\cfill),\\
	\semrole(w',\actionverb),
	\end{array}\right.\\
	&
	\left.\begin{array}{l}\semrole(w_1,\theme),\end{array}\right.\\
	&
	\left.\begin{array}{l}
	\semrole(w_2,\goal),\\
	\isa(\cfill,\cfill), \ldots
	\end{array}\right).
	\end{array}
	\label{eq:filling-roles}
\end{equation}}% 
The distributions show that, conditioned on the semantic role of a 
word, two clearly separable clusters of concepts loom in 
the taxonomy. For the \theme\ role of a filling action, all 
substances/liquids gain considerably high probability, whereas the 
\goal s of such an action are represented by all types of 
containers. Note that also categories \textit{not} explicitly  
modelled, such as \textit{water.n.06},
\textit{soup.n.01}, or \textit{spoon.n.01} and \textit 
{bowl.n.03} and \textit{glass.n.02}, respectively, have been assigned 
significant probability masses indicating that the model indeed 
resonably generalizes across object categories.

%In a WSD problem, interest is
%less focused on computing a posterior distribution over \textit 
%{all} categories but rather to compute the most probable 
%meaning of a particular word out of a selection of senses. 
%From a probabilistic point 
%of view, this boils down to supplying a query with additional 
%knowledge about which meanings are certainly inapplicable to the 
%word under consideration. 
%Figure~\ref {fig:adding-amount-bowl} and 
%\ref{fig:filling-goal-bowl} show the posteriors over the meaning of 
%the word `bowl' in context of the two sentences `add a bowl of 
%water' and `fill a bowl with water'. In the former case, the term 
%`bowl' refers to an abstract entity, a unit of measure specifying 
%the amount of how much water is to be added, whereas in the latter 
%case it determines a physical object, the container to be filled 
%with water. As can be seen, although neither of the two concepts 
%have been seen in the training data, the learned MLN selects 
%the word meanings which are most plausible in the given context.

\begin{figure*}[t]
	%\begin{subfigure}[b]{.40\textwidth}
	\centering
	\begin{tabular}{ll}
		\includegraphics[width=.4\textwidth]{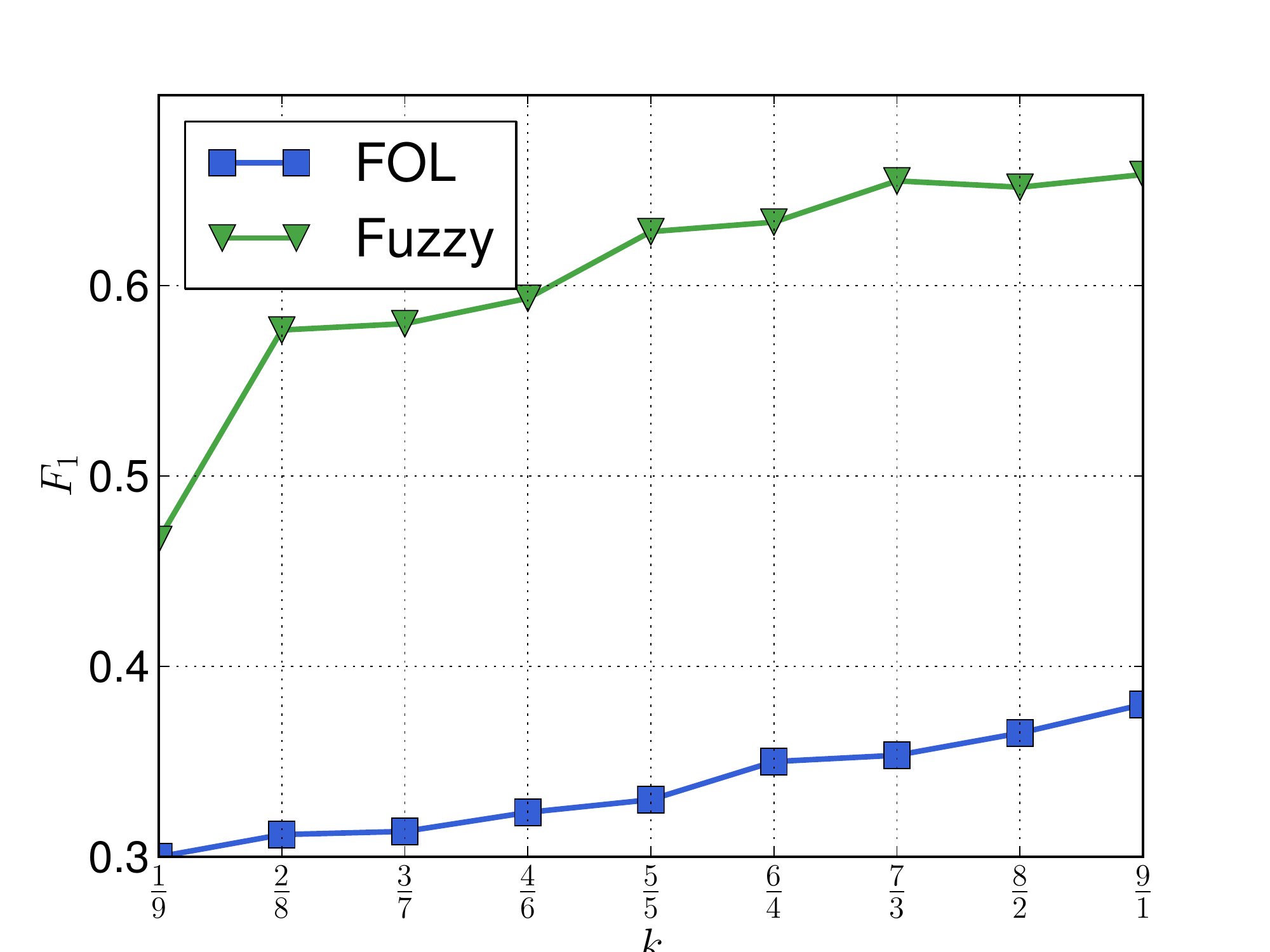} &
		%\caption{}
		%\label{fig:fol-vs-fuzzy-avg}
	%\end{subfigure}
	%\begin{subfigure}[b]{.60\textwidth}
	%\vspace{-5ex}
	    \begin{minipage}{.55\textwidth}
	    \vspace{-31ex}
                \resizebox{\textwidth}{!}{
                \begin{tabular}{|ll|l|l|l|l|l|l|l|l|l|}
                        \hline
                        \multicolumn{2}{|l|}{\multirow{2}{*}{\textbf{Action Verb} }}              & \multicolumn{9}{c|}{$k$} \\
                                          &                         & $\nicefrac{1}{9}$ & $\nicefrac{2}{8}$ & $\nicefrac{3}{7}$  & $\nicefrac{4}{6}$ & $\nicefrac{5}{5}$ & $\nicefrac{6}{4}$ & $\nicefrac{7}{3}$ & $\nicefrac{8}{2}$ & $\nicefrac{9}{1}$ \\ \hline
                        \multirow{2}{*}{\textit{Filling}}   & \scriptsize{FOL} &      0.40 & 0.41 & 0.41 & 0.42 & 0.44 & 0.49 & 0.44 & 0.46 & 0.51 \\\cline{2-11}
                                                    &\scriptsize{Fuzzy}&      \textbf{0.64} & \textbf{0.69} & \textbf{0.67} & \textbf{0.68} & \textbf{0.75} & \textbf{0.75} & \textbf{0.75} & \textbf{0.75} & \textbf{0.75} \\\hline
                        \multirow{2}{*}{\textit{Adding}}    & \scriptsize{FOL} &      0.27 & 0.29 & 0.29 & 0.32 & 0.29 & 0.34 & 0.38 & 0.36 & 0.38 \\\cline{2-11}
                                                    &\scriptsize{Fuzzy}&      \textbf{0.44} & \textbf{0.50} & \textbf{0.49} & \textbf{0.51} & \textbf{0.49} & \textbf{0.52} & \textbf{0.57} & \textbf{0.56} & \textbf{0.57} \\\hline
                        \multirow{2}{*}{\textit{Slicing}}   & \scriptsize{FOL} &       0.28 & 0.30 & 0.30 & 0.31 & 0.31 & 0.34 & 0.34 & 0.34 & 0.34 \\ \cline{2-11}
                                                    &\scriptsize{Fuzzy}&       \textbf{0.36} & \textbf{0.49} & \textbf{0.54} & \textbf{0.48} & \textbf{0.60} & \textbf{0.56} & \textbf{0.61} & \textbf{0.61} & \textbf{0.65} \\\hline
                        \multirow{2}{*}{\textit{Cutting}}   & \scriptsize{FOL} &       0.27 & 0.28 & 0.29 & 0.29 & 0.32 & 0.32 & 0.34 & 0.34 & 0.34 \\\cline{2-11}
                                                   &\scriptsize{Fuzzy}&       \textbf{0.40} & \textbf{0.51} & \textbf{0.57} & \textbf{0.62} & \textbf{0.64} & \textbf{0.64} & \textbf{0.66} & \textbf{0.66} & \textbf{0.66} \\\hline
                        \multirow{2}{*}{\textit{Putting}}  & \scriptsize{FOL} &      0.42 & 0.43 & 0.43 & 0.44 & 0.46 & 0.45 & 0.46 & \textbf{0.53} & \textbf{0.55} \\\cline{2-11}
                                                    &\scriptsize{Fuzzy}&      \textbf{0.43} & \textbf{0.48} & \textbf{0.48} & \textbf{0.50} & \textbf{0.53} & \textbf{0.50} & \textbf{0.51} & 0.51 & 0.50 \\\hline
                        \multirow{2}{*}{\textit{Stirring}}   & \scriptsize{FOL} &      0.16 & 0.16 & 0.16 & 0.16 & 0.16 & 0.16 & 0.16 & 0.16 & 0.16 \\\cline{2-11}
                                                   &\scriptsize{Fuzzy}&      \textbf{0.53} & \textbf{0.79} & \textbf{0.73} & \textbf{0.77} & \textbf{0.76} & \textbf{0.83} & \textbf{0.83} & \textbf{0.82} & \textbf{0.82} \\\hline
                \end{tabular}}
		\end{minipage}
	\end{tabular}
		%\caption{}
		%\label{fig:fol-vs-fuzzy-results}
	%\end{subfigure}
	\vspace{-3ex}
	\caption{\textit{Left:} $F_1$ scores averaged over all action verbs. \textit{Right:} $F_1$ scores for inverse $k$-fold cross validation for $k=\nicefrac{1}{9}\ldots\nicefrac{9}{1}$.}
	\label{fig:fol-vs-fuzzy-table}
\end{figure*}
\section{Experiments}
\label{sec:experiments}

%For evaluating the performance of \fuzzymln s, we chose the 
%domain of word sense disambiguation, which is a common and widely 
%studied problem in the discipline of natural-language processing. 
We evaluate our method by comparing its performance against 
classical MLNs with FOL semantics being applied 
to the problem of word sense disambiguation. We use a real-world 
data set of natural-language instructions that have been mined 
from the wikihow.com web site and manually annotated with
correct word senses. We take into account sense co-occurrences and 
part-of-speech tags. The MLN thus only contains one single template 
formula, 
{\footnotesize
\begin{align*}
        &\phantom{\land\ \ \ }\pos(w_1,+p_1)\land\pos(w_2,+p_2)\\
	&\land\ \hassense(w_1,s_1)\land \isa(s_1,+c_1)\\ 
	&\land\ \hassense(w_2,s_2)\land \isa(s_2,+c_2)\ 
	%&&\land\ \semrole(?w_1,+?r_1)\land\semrole(?w_2,+?r_2)\\
	\land\ w_1\neq\ w_2.
\end{align*}}%
\noindent In order to showcase the generalization capabilities of \fuzzymln s, 
we chose the hardest experimental setup we can imagine: 
(1) the datasets have been selected to exhibit maximal entropy with 
respect to the concepts that are contained in the examples, so that 
they are as dissimilar as possible, and (2) the model was trained 
with only very small portions of training data. We conduct 
`inverse' $k$-fold cross-validation, a modification of traditional 
cross-validation, where also inverse proportions of training and 
test set sizes are considered. For $k=\nicefrac{1}{9}$ , for 
example, we use only 10\% of the data available for training the 
model, and the remaining 90\% serve for evaluation. Conversely, 
$k=\nicefrac{9}{1}$ corresponds to classical 10-fold cross 
validation.

We group the instructions with respect to the action verbs they 
contain and use 20 examples per action verb in each fold. The 
results are shown in Figure~\ref{fig:fol-vs-fuzzy} and \ref 
{fig:fol-vs-fuzzy-table}. \fuzzymln s clearly outperform the 
classical MLNs in almost \textit{every} test case. Moreover, 
\fuzzymln s achieve $F_1$ scores significantly above 0.5 even with 
very small portions of training data (cmp. `filling' with only 10\% 
of the data). The $F_1$ score measures the classification accuracy 
wrt. word meanings from the taxonomy assigned to each word in the 
respective NL instruction. It is interesting to note that, while 
only moderate improvements in classical MLNs are recorded with 
increasing amounts of training databases, the most significant 
performance jumps with \fuzzymln s can be observed when only 
sparse training data is used. In these extreme cases, where 
concepts occur in the test data that are not contained in the 
training data, classical MLNs (and all other approaches mentioned 
in the related work) are inapplicable to perform 
meaningful reasoning but are forced to randomly guess. This shows 
that fuzzy inference in MLNs can perform adequate reasoning about 
concepts in the taxonomy that are not explicitly represented in the 
probability distribution and have not been seen during training.

        %\item `Filling': Fuzzy reaches its maximal performance (0.75) already with half of the data,
        %whilst FOL is only at 0.50 with 90\% data
        %\item Error cases: Bugs in the WordNet taxonomy, e.g.\ \cpanone\ and \cpanthree\
        %in WordNet are both containers, but only \cpanthree\ inherits from \ccontainer.
        %The model therefore prefers \cpanthree\ though \cpanone\ is labelled.
%\end{itemize}

%\section{Evaluation}
%
%We evaluate our formalism by applying it to the problem of word-sense
%disambiguation on two benchmark datasets and one dataset of 
%natural-language instructions. 
%
%We will answer the following questions:
%\begin{enumerate}
	%\item How far can we push the size of the training set for the
	%MLN still computationally tractable.
	%\item How far can we shrink the training set for the MLN to still
	%generalize well in off-domain applications.
%\end{enumerate}
\section{Related Work}

A couple of frameworks have been proposed to incorporate concept 
taxonomies and similarity in probabilistic models, such as 
probabilistic description logics~\cite{Lukasiewicz2008852,niepert2011log} (PDL), 
tractable Markov logic (TML)~\cite{domingos12tractable} and 
probabilistic similarity logic (PSL) \cite {brocheler2012psl}, 
which differ from \fuzzymln s in basically two fundamental ways: (1) 
\fuzzymln s do \emph{not} postulate uncertainty among the 
taxonomy structure as such, i.e. the structure itself is not 
subject to reasoning and (2) \fuzzymln s do not model the whole 
taxonomy in the probabilistic model, but only the concepts seen 
during training. This makes \fuzzymln s a more compact 
reasoning framework. TML is a subset of Markov logic 
networks. TML introduces the idea of concept taxonomies in MLNs, 
but in order to perform reasoning about superclasses, the 
inheritance relationship of concepts is explicitly represented in 
the model. By employing semantic similarity as evidence, the 
taxonomy relation is more compactly encoded in \fuzzymln s. PSL 
uses a formalism similar to \fuzzymln s. Unlike \fuzzymln s, 
however, the goal of PSL is rather to reason about  
degree to which a set of entities are similar to each other. Conversely, in \fuzzymln s 
the taxonomy is fixed and serves for filling gaps in the 
probabilistic KB. Hybrid MLNs (HMLN)~\cite 
{wang08hybrid} extend MLNs to reason about continuous variables. 
They discern features in hard FOL and 
numeric features that may be expressed as `soft' (in)equality 
constraints. Those constraints are 
typically connected in a multiplicative way, such that, if a 
logical constraint evaluates to false, then also a connected 
numeric feature will have no influence on the probability of the 
respective possible world. Hence representing semantic similarities 
in HMLNs does not appear straightforward.  The concept of soft 
evidence~ \cite {jain10soft} is closely related to the idea of 
vague evidence, though it has fundamentally different semantics for 
it still assumes boolean truth values and soft evidences 
serve as prior probability constraints on ground atoms. 

To the best of our knowledge, none of these approaches can 
deal with entities that are not part of the probabilistic model in 
any meaningful way. This is a severe limitation, because they are 
not capable of exhaustively modelling joint probability distributions 
of realistic domain sizes. Since learning in first-order 
probabilistic models remains intractable in the general case, 
inference and generalization across concepts is 
essential and outstandingly important for probabilistic 
relational models to be scalable and applicable to real-world 
problems.

\vspace{-1ex}
\section{Conclusions} 

In this work, we have described the design and the implementation of
\fuzzymln s, an extension of MLNs that allows us to represent
probability distributions over open domains compactly -- if complete
ontologies are available for these domains. The basic idea underlying
\fuzzymln s is to explicitly represent only the small subset of
concepts that is contained in the training databases. After having
learned the probability distribution \fuzzymln s can reason about
concepts that are not contained in the graphical model but in the
taxonomy. They do so by exploiting the fact that the relational structure of
concepts in the taxonomy is correlated with the relational structures
of the explicitly represented concepts weighted by a notion of
semantic similarity. \fuzzymln s implement this bias by generalizing
the \isa\ assertions for off-domain concepts from boolean truth to
real-valued degrees of truth. The degree of truth is then computed
based on the semantic similarity of the off-domain
concept to those concepts contained in the graphical model. 

We have shown that \fuzzymln s can perform different probabilistic 
reasoning tasks in a way that matches our intuitions and can
outperform probability distributions learned in the ordinary MLN
framework both significantly and substantially.

%----------------------------------------------------------------------

\section{Acknlowlegments}
\label{sec:ack}

This work has been supported by the EU FP7 projects \textit{RoboHow} (grant no. 288533)
and \textit{ACAT} (grant no. 600578).

%----------------------------------------------------------------------

\bibliographystyle{aaai}
%\allbibliography{literature}

\end{document}